\documentclass[conference,twocolumn]{IEEEtran}
\usepackage{graphicx}
\usepackage{algorithm}
\usepackage{algorithmic}
\usepackage{listings}
\usepackage{authblk}
\usepackage{amsmath} 
\usepackage{etoolbox}
\makeatletter
\patchcmd{\@makecaption}
  {\scshape}
  {}
  {}
  {}
\makeatother

\begin{document}

\title{Use of a genetic algorithm to find solutions to introductory physics problems}

\author[1]{Tom Bensky}
\author[2]{Justin Kopcinski}
\affil[1]{California State University San Luis Obispo\\Department of Physics, San Luis Obispo, California, USA \authorcr Email: {\tt{tbensky}@calpoly.edu}\vspace{1.5ex}}
\affil[2]{Department of Computer Engineering \authorcr Email: {\tt{jkopcin}@calpoly.edu}\vspace{1.5ex}}


\maketitle

\begin{abstract}
In this work, we show how a genetic algorithm (GA) can be used to find step-by-step solutions to introductory physics problems.  Our perspective is that the underlying task for this is one of finding a sequence of equations that will lead to the needed answer. Here a GA is used to find an appropriate equation sequence by minimizing a fitness function that measures the difference between the number of unknowns versus knowns in a set of equations.  Information about knowns comes from the GA posing questions to the student about what quantities exist in the text of their problem.  The questions are generated from enumerations pulled from the chromosomes that drive the GA.  Equations with smaller known vs. unknown differences are considered more fit and are used to produce intermediate results that feed less fit equations.   We show that this technique can guide a student to an answer to any introductory physics problem involving one-dimensional kinematics.  Interpretability findings are discussed.
\\
\\
{Keywords: genetic algorithm, physics problem solving, socratic tutorial, automated tutorials}
\end{abstract}

\section{Introduction}

The development of an automated ``general problem solver'' for use by students needing help with introductory physics problems has been elusive over the past few decades\cite{ai_01}\cite{ai_02}\cite{ai_03}\cite{bensky}.  Instead, countless animations, electronic supplements, and numerical simulations to textbook problems have appeared\cite{phet}\cite{compadre}\cite{mastering}, some flourishing\cite{phet_global}. Nonetheless, no technology has emerged that can incrementally guide a student to the solution to their particular problem. 

Large Language Models (LLMs) continue to show great potential in this area\cite{ref02}\cite{phi4}, but their probabilistic nature make it difficult to always expect thoroughly proper results, with today's models producing correct solutions ${\sim}40-70\%$ of the time\cite{ref02}\cite{llm_physics1}. Additionally, as impressive as LLM responses are to physics problem prompts, they may not be the most pedagogically effective, as they only provide the student with a solution without any meaningful interaction\cite{phys_teach_llm}.  In particular, LLMs don't involve student input or ask follow-up questions during their reasoning process. OpenAI's recent ``study mode''\cite{openai_study_mode} may help in this regard, but its effectiveness remains to be seen.

In this work, we show how finding a solution to a physics problem can be seen as a process of finding an optimal sequence of equation use that leads to a solution. Our motivation for using an optimization technique for this (and a genetic algorithm in particular) is threefold.

First, inspecting a publisher's ``solutions manual'' to a typical introductory physics textbook, one will see that arriving at the solution to a problem is generally possible by applying a small set of relevant equations in the proper order\cite{solns_manual}.  The notion of a proper ordering of equations implies the existence of some non-obvious optimization that somehow leads to an answer.

Second, likely the optimal learning mode for any student is a one-on-one interaction with an expert (such as a teacher), where the student can get help with their particular problem. In such an interaction, the expert would ask the student short and specific questions about their work, in a way that guides them with incremental progress toward an answer\cite{socratic}.   Here, we use answers to these questions as inputs that guide the optimizer as it runs. 

Third, there are a few characteristics of a general problem solver we demand. These include allowing for different approaches to identical problems, strong interaction with the student in a question-and-answer dialog, and flexibility to accommodate a variety of student input. Lastly, the algorithm cannot be overly greedy, brittle, or difficult to extend in order to handle more complex problems.  These were all shortcomings of previous work\cite{bensky}.

We find our demands above to be naturally met by a genetic algorithm (GA) since they typically start  with a random set of chromosomes\cite{koza}\cite{michalewicz} that become more refined as it runs (much like our expectation of a solution to a physics problem). We will show how interactions with the student can come from the instantaneous state of the chromosomes. The GA will have a fitness function linked to how well inputs from the problem correspond to unknowns variables in a domain of relevant equations. We also note that evolution of a GA does not aggressively pursue any particular path toward a solution, and is able to adapt as it goes.

Lastly, we are not developing a solver that does natural language processing or semantic understanding of a problem\cite{ai_03}. Additionally, this work does not provide numerical calculations, do symbolic algebra, or numerically simulate a problem with animations or otherwise. All such tasks are left up to the student.  The goal of this work only provides textual guidance to the student about their problem.

\section{Example}
\label{sect:example}

As an example, suppose an expert is working with the student on the following problem:

\begin{quote}
{\sl
A car at rest accelerates from a stop light at 5 m/s$^2$ for 8 seconds, then coasts for 3 more seconds.  How far has it traveled?
}
\end{quote}

Here an expert may begin by asking the student ``Do you see any objects in the problem?'' for which the student may respond ``A car.'' Next, the expert may ask ``Do you know anything about the position of the car?'' for which the student may response ``It starts at a stoplight.'' A further question may lead to the initial speed of the car (which is zero), and so on.

This is the kind of interaction we seek for the student: small incremental progress toward a solution, that forces them to consider and reconsider the text of their problem.  

We note that three critical pieces of information have become known. The first is a subscript of ``car'' for use in an equation like 
\begin{equation}
\label{xcar}
x_{car}=x_{0,car} +v_{0x,car}\Delta t+\frac{1}{2}a_{x,car}\Delta t^2,
\end{equation}
giving an equation dedicated to the position of just the car.  Second if the initial position of the car or $x_{0,car}=0$ is known,  this equation would become
\begin{equation}
x_{car}=v_{0x,car}\Delta t+\frac{1}{2}a_{car}\Delta t^2.
\end{equation}
Lastly, that the car is known to start at rest meaning $v_{0x,\textrm{car}}=0$ giving
\begin{equation}
x_{car}=\frac{1}{2}a_{car}\Delta t^2.
\end{equation}

As one can see, a sequence has begun that has taken an equation with five unknowns and reduced it to three.  The expert might now carefully watch this equation as the dialog continues: should two of any of the three variables, $x_{car}$, $\Delta t$, or $a_{car}$ become known, the third may be solved algebraically, which will become a new known quantity about the problem, that may aid subsequent steps.  In this case, asking if a time interval is known that also corresponds to a given acceleration, will yield $\Delta t=8$ seconds and $a=5$ m/s$^2$, allowing one to compute $x_{car}$, which is a critical step in solving this problem.

The premise of the work here is to have the algorithm ask the student a set of simple (yes/no-style) questions, forcing them to take repeated looks at their problem.  With their answers, the GA will work to order a known set of physics equations, leading to an answer by minimizing the unknowns amongst the equation set. Eventually, a final unknown can be solved for that will yield the answer they seek.

We now discuss the origin of the questions and the domain that will drive the algorithm.

\section{The genetic algorithm optimization}

\subsection{Domain}

To begin, we limit the work here to one-dimensional kinematic problems with constant acceleration, involving one or more objects. We claim that all such problems can be solved using some sequence of the three equations shown in Table~\ref{table_equation_domain}, that shows the usual equations of introductory kinematics (with some additional labels).  

One label is the equation number. Since information in a chromosome is generally enumerated, equations will always be referenced by their number.  The other label is a variable enumeration within each equation. For example from this table, the tuple $(1,2)$ would refer to equation 1 and variable 2 or $x_0$. Likewise $(2,3)$ would refer to $a_x$.

\begin{center}
\begin{table}[h]
\centering
\begin{tabular}{ c | c |  c }
\hline
Equation\\Number & Equation & Unknowns\\
\hline\hline
1 & $\stackrel{\textcircled{1}}{x}=\stackrel{\textcircled{2}}{x_0}+\stackrel{\textcircled{3}}{v_{0x}}\stackrel{\textcircled{4}}{\Delta t}+\frac{1}{2}\stackrel{\textcircled{5}}{a_x}\Delta t^2$ & 5 \\
2 & $\stackrel{\textcircled{1}}{v_x}=\stackrel{\textcircled{2}}{v_{0x}}+\stackrel{\textcircled{3}}{a_x}\stackrel{\textcircled{4}}{\Delta t}$ & 4 \\
3 & $\stackrel{\textcircled{1}}{\Delta t} = \stackrel{\textcircled{2}}{t}-\stackrel{\textcircled{3}}{t_0}$ & 3 \\   
\\
\end{tabular}
\caption{The domain of equations used in this work. Each equation is characterized by its own number, the textual representation of the equation itself, the number of unknowns in the equation, and a unique enumeration for each variable in the equation (circled numbers). One such table exists for each object and per each acceleration zone in the problem.}
\label{table_equation_domain}
\end{table}
\end{center}
For the work here, this tuple must be expanded with two additional elements.  The first is needed because each object in a problem (i.e. a car, truck, etc.) requires its own table of equations unique to it.  Thus referencing Table~\ref{table_equation_domain} would be of the form $(n,1,2)$ for example, to reference variable $2$ of equation $1$ for object $n$ (with $n=0$ for the car and $n=1$ for the truck for instance).  

Second, each object requires such a table for each acceleration zone it experiences in a problem. Here, an ``acceleration zone'' is a segment of the problem where the object $n$ experiences a particular acceleration.  Thus, the complete tuple is of the form $(n,e,v,z)$ where $n$ is the object number, $e$ the equation number,  $v$ the variable number (found in equation $e$), and $z$ is the acceleration zone. Table~\ref{table_equation_domain} can be thought of as needing to be subscripted with $n$ (the object as in Eqn.~\ref{xcar}) and $z$ for full and proper use here.

\subsection{Chromosome format}
 
The four-dimensional tuple of the form $(n,e,v,z)$ described above sets our plan for interpreting and enumerating the chromosomes.  For each element of the tuple, we use a three-bit binary string, allowing a maximum decimal value of $7$ (or $8$ unique binary combinations) for each element (which is a hyper-parameter of this work).  With this setting, we can have at most eight: objects, equations per object, variables per equation and acceleration zones per object. We find this adequate for solving most introductory problems.  

We use a chromosome structure consisting of many groups of lengths 12 as in
\begin{equation}
b_1b_2b_3 b_4 b_5 b_6 b_7 b_8 b_9 b_{10}b_{11}b_{12},
\label{eqn:chrome}
\end{equation}
where $b_n\in\{0,1\}$ and $b_1b_2b_3$ the binary encoding for the object number, $b_4b_5b_6$ the equation number,  $b_7b_8b_9$ the variable number, and  $b_{10}b_{11} b_{12}$ the acceleration zone. Thus a tuple can be constructed as 
\begin{equation}
(\textrm{dec}(b_1b_2b_3),\textrm{dec}(b_4b_5b_6),\textrm{dec}(b_7b_8b_9),\textrm{dec}(b_{10}b_{11}b_{12})),
\end{equation}
where $\textrm{dec}$ is a function that converts a binary string into a decimal equivalent. 

In summary, an individual chromosome consists of multiple $12$-bit binary groups from which the tuples are derived. With trial and error, the work presented here uses chromosomes with a length of $12,000$ bits, meaning they contain $1,000$ $12$-bit groups\cite{coupon}. The GA is seeded by generating an initial population of $50$ such chromosomes, each consisting of a uniformly random distribution of 1s and 0s.

\subsection{Generating the questions}
\subsubsection{Starting out}

With the enumeration of the chromosomes defined, we now show how questions for the student will be generated.  Suppose that a chromosome exists in the following format above as (dots and bars for clarity)
\begin{equation}
\vert 001\cdot010\cdot 010\cdot 001\vert 001\cdot 001\cdot 010\cdot 001\vert 001\cdot 011\cdot 010\cdot 010\vert\dots
\end{equation}
Again using the unit of four, three-bits groups, this translates to decimal values of
\begin{equation}
1221\cdot 1121\cdot  1322.
\end{equation} 
Thus this chromosome is focusing on 1221, which means object 1, equation 2, variable 2, and acceleration zone 1 (with the 1121 and 1322 translating similarly). Questions for the student to answer are generated as follows.

A data structure holding a list of objects seen in the problem is maintained, which is initially empty.  Guided by the 1221, we look for object 1 (which initially is non-existent), so the question ``Do you see any objects in the problem?'' is asked. The student response (likely ``a car'' will go into this list as object 1). As a Python dictionary, this would resemble
\begin{verbatim}
objects = {1: 'a car'}.
\end{verbatim}

\begin{figure}[t]
\centering
\includegraphics[scale=0.8,width=\columnwidth]{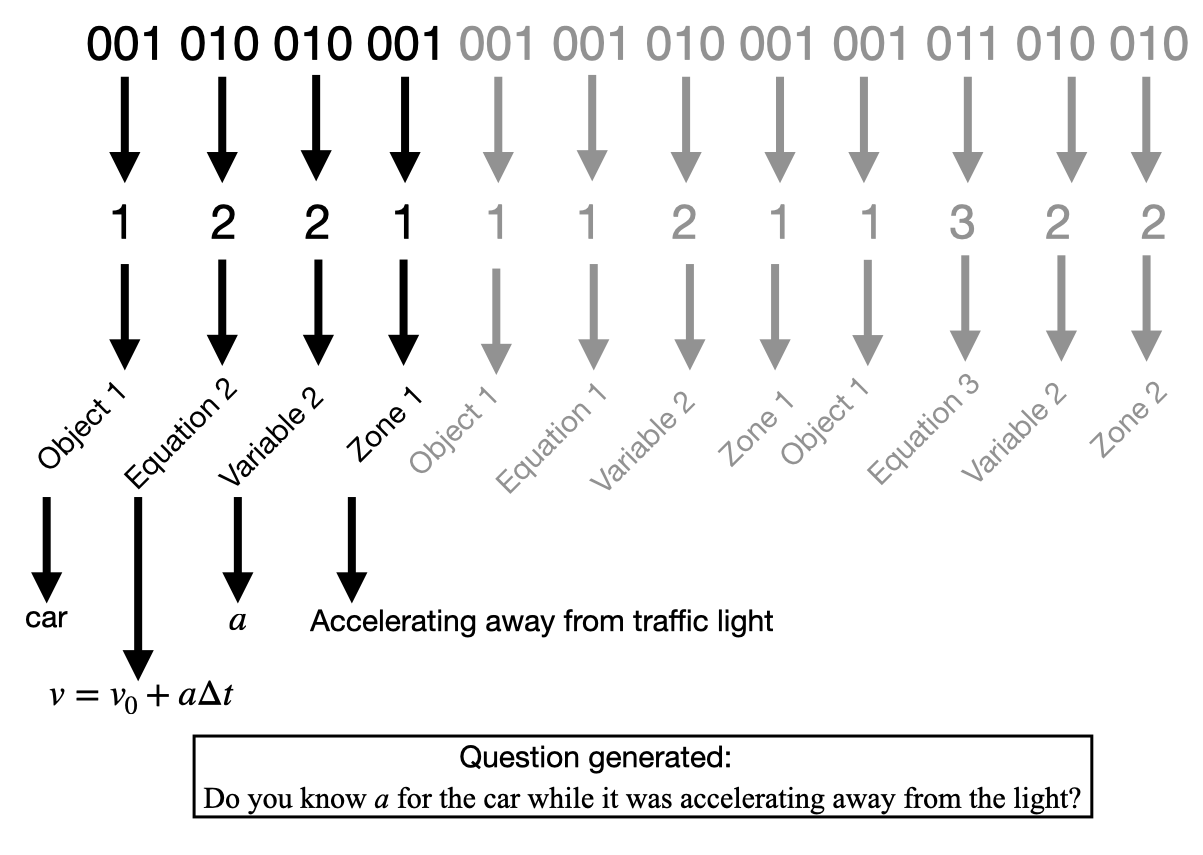}
\caption{Decoding a chromosome into a question for the student. Two more questions would follow (not shown) from the subsequent values.}
\label{fig:decode}
\end{figure}

Continuing to interpret this chromosome (for object 1, now known to be ``a car''), equation 2 and variable 2 is referenced, so the question ``Do you know $a_x$ of the car?'' is asked. In short, the student is being asked if they know some acceleration of the car. Forming questions from the chromosomes is illustrated in Fig.~\ref{fig:decode}.

Suppose the student answers ``5 m/s$^2$'' in response.  With $a_x$ of the car now known, our plan of minimizing unknowns in the equation set is proceeding since equation 2 now has only 3 unknowns (down from its initial 4). The last value in the tuple about the acceleration zone of the car however, must also be resolved.  

A data structure storing information on acceleration zones is now consulted about acceleration zone 1 (the final ``1'' in 1221). Since nothing is initially known about any acceleration zones, the student is asked ``What was the car doing when it was given the acceleration which you said was accelerating at 5 m/s$^2$?'' The student must peer back at the problem and reply with something like ``Accelerating away from the traffic light,'' which becomes the description stored about acceleration zone 1 as follows
\begin{verbatim}
zones = {1: "Accelerating away from 
                the traffic light"}.
\end{verbatim}

A data structure called \verb!knowns! is now populated with this information and will resemble
\begin{verbatim}
        knowns = [
             {
                  'object': 1, 
                  'eqn': 2, 
                  'var': 2, 
                  'zone': 1,
                  'response': '5 m/s^2'
             }
            ].
\end{verbatim}

In sum, parsing of a chromosome for questions is a logical and well defined process.  Answers from the student will fill the data structures \verb!objects!, \verb!knowns!, and \verb!zones!, as the solution search proceeds.

\subsubsection{Pedagogical caution}
\label{Pedagogical caution}

Given that this algorithm is a pedagogical tool, some caution must be imposed on the logic. This caution is that non-obvious interrelationships exits between all variables in a given equation that must be observed, given the accumulation of past knowns by the question asking logic. The interrelationships are as follows.

Consider equation 2 in Table~\ref{table_equation_domain}, $v_x=v_{0x}+a_x\Delta t$. The most obvious relationship between $v_x$ and $v_{0x}$ is that $v_x$ the speed an object eventually attains after starting out with speed $v_{0x}$ and having an acceleration $a_x$ applied to it for a time interval of $\Delta t$. More subtle relationships also hold, for example between $a$ and $v_{0x}$: $a$ is the acceleration an object was given at the moment it was started out with a speed of $v_{0x}$.

Thus, the question generating logic must provide warnings to the student before accepting their responses at face value, as to what they claim know. We illustrate this by going on to the next tuple in the chromosome, which is $1121$. This asks about object 1 (now known to be ``the car''), and equation 1, variable 2 ($x_0$) in acceleration zone 1 (now known to be ``Accelerating toward the traffic light'').

The cautionary question generated will be complex, given knowledge acquired from the previous question, but is still straightforward to generate. In this case,  ``Do you know the initial position for {\bf the car}?'' will not be sufficient given the history of responses. We must instead ask  ``Do you know the initial position for {\bf the car} at the start of {\bf its acceleration away from the traffic light}, which should be the same instant when the car had $a_x$, which you said was {\bf 5 m/s$^2$}?'' Note that bold words came from the student's response to a previous question.

These cautionary questions do not come from the chromosomes. They all fit nicely into a single template (internal to the code) which is
\begin{quotation}
 {\sl 
 Do you know the $<$variable description$>$ of the $<$object$>$ at the $<$start of$><$end of$>$ when it was $<$acceleration-zone description$>$?
 }
\end{quotation}

Here the bracketed strings are tags to be replaced with contextual information at the instant the caution message is needed.  The tag $<$variable description$>$ is the variable the student has claimed to know, $<$object$>$ is the object in the problem under scrutiny, one of $<$start of$><$end of$>$ is chosen as linked to a initial or final variable (i.e. $x_0$ vs. $x$), and $<$acceleration-zone description$>$ is a previous student-given description of they saw the object doing.

All told, 56 such interrelationships exist amongst the three equations in Table~\ref{table_equation_domain}. They are defined primarily by which new variable the student is now claiming to know  given some past variable they claimed to have known.  Pedagogically sound descriptions of the relationships between any two variables for the student to ponder are important. A partial list is shown in Table~\ref{table_warnings}.

\begin{center}
\begin{table}[b]
\centering
\begin{tabular}{ c | c |  l }
\hline
New & Past & Relationship\\
\hline\hline
$v_{0x}$ & $x$ & $v_{0x}$ is the velocity before ending up at position $x$\\
$v_{0x}$ & $x_0$ & $v_{0x}$ is the speed at the instant it was a position $x_0$\\
$v_{0x}$ & $\Delta t$ & $v_{0x}$ is the speed at the beginning of interval $\Delta t$\\
$v_{0x}$ & $a_x$ & $v_{0x}$ is the speed as it obtained acceleration $a_x$\\
$v_{0x}$ & $v_x$ & $v_{0x}$ is the speed before it got to speed $v_x$\\
$v_{0x}$ & $t$ & $v_{0x}$ is the speed before time ticked up to $t$\\
$v_{0x}$ & $t_0$ & $v_{0x}$ is the speed of at the instant of time $t_0$\\
\dots & \dots & \dots\\
\dots & \dots & \dots\\
\\
\end{tabular}
\caption{A sample of the 56 interrelationships that exist between the variables in Table~\ref{table_equation_domain}. These are the relationships that exist between $v_{0x}$ and the other variables in $v_x=v_{0x}+a_x\Delta t$. In practice, these are embellished with more pedagogical aides.}
\label{table_warnings}
\end{table}
\end{center}

\subsubsection{Finding new knowledge}

As knowledge of variables accumulate, the algorithm constantly polls the domain of equations, looking for cases where the number of knowns in a given equation equals the number of unknowns minus one.  This is a case where the missing unknown variable can be solved for algebraically. 

In such a case text is output to the student informing them of this possibility and the newly found variable is automatically inserted into the \verb!knowns! structure, with no question needed.  A textual description is also tied to the new unknown, indicating that it was {\sl solved for} (i.e. not pulled from the problem directly).  As per the spirit of this work, the student is left to do the algebra to actually solve for the unknown variable when advised to do so.

\subsubsection{Efficiency and required student effort}

As mentioned, each chromosome consists of encodings for $1,000$ tuples, which could lead to potentially $1,000$ generated questions per chromosome to answer, which not practical. This potentially large number of questions can be dramatically reduced using a few somewhat obvious mechanisms, outlined here. 

\begin{enumerate}

\item The tuple (and any potential question) is ignored if variable $v$ of equation $e$ for object $n$ in acceleration zone $z$ is already known. 

\item We start with $n$, the object number. If this object number is not found in the \verb!objects! dictionary, the user is asked ``Do you see any objects in the problem?.'' Using $3$ bits for this value gives a maximum number of objects per problems to be $8$.  A typical physics problem usually has 1 - 2 objects.  Thus most binary combinations for $n$ pulled from a chromosome will be unneeded, but will still trigger this question. Thus a stop-question is also asked which is ``Do you see any more objects in this problem?'' If the user answers ``no'' then any values of $n$ pulled from chromosome that are not already in the \verb!objects! dictionary are ignored for the remainder of the session.

\item Next, we pull $e$ and $v$ from the chromosome.  The tuple could be ignored for two reasons. The first is if $e$ is greater than the number of equations available. The second is within equation $e$, if the value of $v$ is greater than the number of the equation's unknown variables (see Table~\ref{table_equation_domain}). 

\item If $v$, $e$ and $n$ are known, but not linked to an acceleration zone $z$, the user is asked what ``What was object [n] doing when it had property [v]?''  The answer to this question has two consequences. First, it enumerates a new acceleration zone $z$ and pairs it with a textual description of what the object is doing, as described by the student.  This enumeration avoids a future question should $z$ come up in a subsequent tuple.  Second, it automatically adds an entry to the \verb!knowns! structure as the tuple $(n,e,v,z)$ is now completely known. This will cause future questions coming from such a tuple to be skipped

\item Examining Table~\ref{table_equation_domain}, we see for example that $v_{0x}$ appears in across both equations $1$ and $2$. If a tuple leads to a question about $v_{0x}$ (for a given object $n$) as it pertains to equation $1$, the user's answer is also used to lodge $v_{0x}$ for equation $2$ as well (since physically it is the same $v_{0x}$).  This can avoid a future question about $v_{0x}$  from a tuple specifically directed at $v_{0x}$ for equation $2$.

\item Equations are constantly monitored to see if the number of knowns for a given $(n,e,z)$ equals to the number of unknown (variables) in an equation {\sl minus one}. This will mean a given $(n,e,z)$ equation can be algebraically solved for the remaining unknown, and inserted into the \verb!knowns! structure as a new and fully known $(n,e,v,z)$ tuple. This will avoid a future question about the just solved for variable.

\end{enumerate}
The ordered items above can be thought of as a list of rules {\sl trying to reject a tuple from leading to a question being asked}.

We acknowledge the computational inefficiency of this chromosome structure. Obviously the bits available to enumerate a tuple could be reduced or customized in size or selection for each element of the tuple.  We see this as a tradeoff in two areas.  The first is in the simplicity of the chromosome format. The second (and perhaps more philosophical) is in allowing the GA mechanisms of crossover and mutation (see below) to proceed unfettered, and free from potential \verb!if! statements, modulo operations, or ties to past responses to more tightly bound chromosome values as needed. 

In the sample problem posed above, we were typically asked ${\sim}5$ questions per generation in the initial generations, which declines as the search for a solution continues.  In fact, as a solution nears, a small group of different questions begin to repeat, which are indicative of the student not realizing certain knowns must be available in the text of their problems.  Indeed the student does not do much, relative to the computational work of the GA, as the algorithm rejects most of what is proposed by each generation of the GA.

\subsubsection{Termination}

At some point in parsing the current generation of chromosomes, a new question will not be found and no new algebraic knowledge will be solved for. At such a juncture there is flexibility in the algorithm to perform any organizational tasks not dictated by parsing the chromosome which may also help to find more fit chromosomes. This is also an opportunity to expand the problem solving ability of the algorithm.

In the case of solving one-dimensional kinematics problems, there are two organizational tasks. The first is to ask the student to properly sequence the acceleration zones temporally (as they exist in the problem), and the second is to ask if variables from one acceleration zone may connect to those in a subsequent zone.  

For the problem stated above, the zone where the car is accelerating comes before the zone where the car is coasting.  Also, the position and speed of the car after it accelerates becomes the initial position and speed of car as it enters its coast phase.  Problems involving multiple objects would additionally ask about the possibility of them meeting in space or time.

With such organizational tasks complete, it is time for the genetic algorithm to take a step in its optimization process, and generate the next generation from which to start again with composing questions. This proceeds as follows.

\section{Taking a step with the genetic algorithm}

With question possibilities and organizational steps exhausted, the algorithm must now take a step which will generate the a new population that will hopefully lead to additional questions that will find even more fit chromosomes.  Before taking a step however, a fitness calculation on each chromosome is needed.

\subsection{Fitness calculation}

The fitness of each chromosome is needed by the GA, which is straightforward to calculate as shown in Algorithm~\ref{alg:fitness}, but requires some care.  In particular, the enumerations in a given chromosome may be out of range for the number of known objects, equations, variables, or acceleration zones.  These chromosomes should not merely be ignored (as with the question generator), but actually penalized by the fitness calculation. This is handled as follows.

The fitness of each chromosome in the initial population should be maximal since nothing is known about the problem initially.  This maximal value is found from a ``worst case'' scenario: assume each $(n,e,v,z)$ tuple in a chromosome represents the maximum number of unknowns of any equation in use. For this work, the value is $5$ (the number of unknowns in equation 1).  Thus for a chromosome containing $1,000$ tuples, the initial fitness will be $5\times 1,000=5,000$.

With a populated \verb!knowns! database, the fitness calculation loops through a chromosome, parsing $(n,e,v,z)$ tuples from it.  Then it iterates through the \verb!knowns! summing how many variables are known for the given object, equation and acceleration zone represented by the tuple. The sum is incremented by the success of line \#9, which looks for a $(n,e,z)$ tuple in the \verb!knowns! database.  If such a tuple made it to \verb!knowns!, it must be associated with {\sl some known variable}, the exact name of which is immaterial.

This completed sum is subtracted from the current chromosome fitness, lowering the fitness in proportion to the known variables to which the chromosome maps. This method also penalizes chromosomes that have the $(e,v)$ portions of their tuples out of range.

\begin{algorithm}
\caption{Computing the fitness of a chromosome}
\label{alg:fitness}
\begin{algorithmic}[1]
\STATE $f \leftarrow\textrm{maximum fitness}$
\STATE $\Delta \leftarrow$ length of tuple quanta
\STATE knowns $\leftarrow$ database of all known quantities in problem
\FOR{$i=0$ \TO length(chromosome)-1 {\bf step} $\Delta$}
	\STATE $(n,e,v,z) \leftarrow$  parsed from chromosome at position $i$
	\IF{$e$ and $v$ are valid}
		\STATE C$\leftarrow 0$
		\FOR{$k\in$ knowns}
			\IF{$k[object],k[equation],k[zone]==(n,e,z)$}
				\STATE C$\leftarrow$ C $+1$
			\ENDIF
		\ENDFOR
		\STATE $f\leftarrow f-C$
	\ENDIF
\ENDFOR
\STATE return f
\end{algorithmic}
\end{algorithm}

By minimizing the fitness in this way, we aim to promote chromosomes that evolve toward smaller and smaller numerical differences between the number of unknowns in the definition of an equation, and the number of knowns know about a problem.

\subsection{Genetic algorithm operators}

With the ability to compute chromosome fitness values, the genetic algorithm can now take a step. Here we use the standard implementation discussed elsewhere\cite{koza}\cite{michalewicz}.

Proceeding, the standard cross-over and mutation operators are used\cite{koza}\cite{michalewicz}.  First, the fitness is computed for each chromosome in the generation, then the chromosomes are sorted according to their fitness.   Second, the sorted fitness values are used as their own probability density for use in a ``roulette wheel'' selection method\cite{koza}\cite{michalewicz} to select two individual chromosomes with a probability that is inverse to their fitness. In short, chromosomes with a lower fitness (i.e. more ``fit'' chromosomes) have a higher probability of being selected.  

Third, with a pair of chromosomes selected, two actions occur on the pair.  The first is a crossover, which occurs with a probability of $25\%$ as recommended\cite{goldberg}. Here, a random index point is selected between $0$ and the length of a chromosome.  All bits between this point and the end of the chromosome are swapped between the pair.  Next, is a mutation where the bits of the two new chromosomes (that resulted from the crossover) are visited one-by-one, and bit-flipped ($1\rightarrow 0$ and $0\rightarrow 1$) with a probability of $0.01$\cite{norvig}.  

The two new chromosomes are saved as the first two in a new generation. This process is repeated until $1,000$ new chromosomes are generated, creating the next generation. These are sent back to the question generation process described above. 

\section{Experiment}

We have implemented the ideas illustrated above in Python, and demonstrate its operation with a session focused on the sample problem shown in Section~\ref{sect:example}.

As mentioned, the GA was seeded with an initial population of $50$ chromosomes, each consisting of $12,000$ random 1s and 0s (or 1000, 12-bit tuples). We implemented a small Python GA library by interpreting considerations for assembling an initial population and handling mutation and crossover as explained in Refs.~\cite{koza}\cite{michalewicz} and \cite{goldberg}. The fitness calculation was coded from the considerations described above.  A detailed summary of our code that drives the GA is given in the Appendix.

When run, Fig.~\ref{fig:sample_run} shows the partial dialog of the session. Here, the bold text was typed from the user and the plain text was generated by parsing the chromosomes as discussed.  This dialog is similar to that shown in Ref.~\cite{bensky}.

Eventually all possible questions parsed from the initial generation were asked, at which time the algorithm took a step. New question possibilities arose, since the mutation and crossover operations will generally alter the chromosomes.  Of note here is that the algorithm found a way of computing the position of the car after the initial acceleration phase from quantities ($a_x$, $v_{0x}$, $\Delta t$ and $x_0$) from equation 1 in Table~\ref{table_equation_domain}, which was added to the list of knowns. This was a case of the algorithm recognizing $N-1$ knowns for an equation with $N$ unknowns.  

\begin{figure}[t]
\centering
\includegraphics[width=\columnwidth]{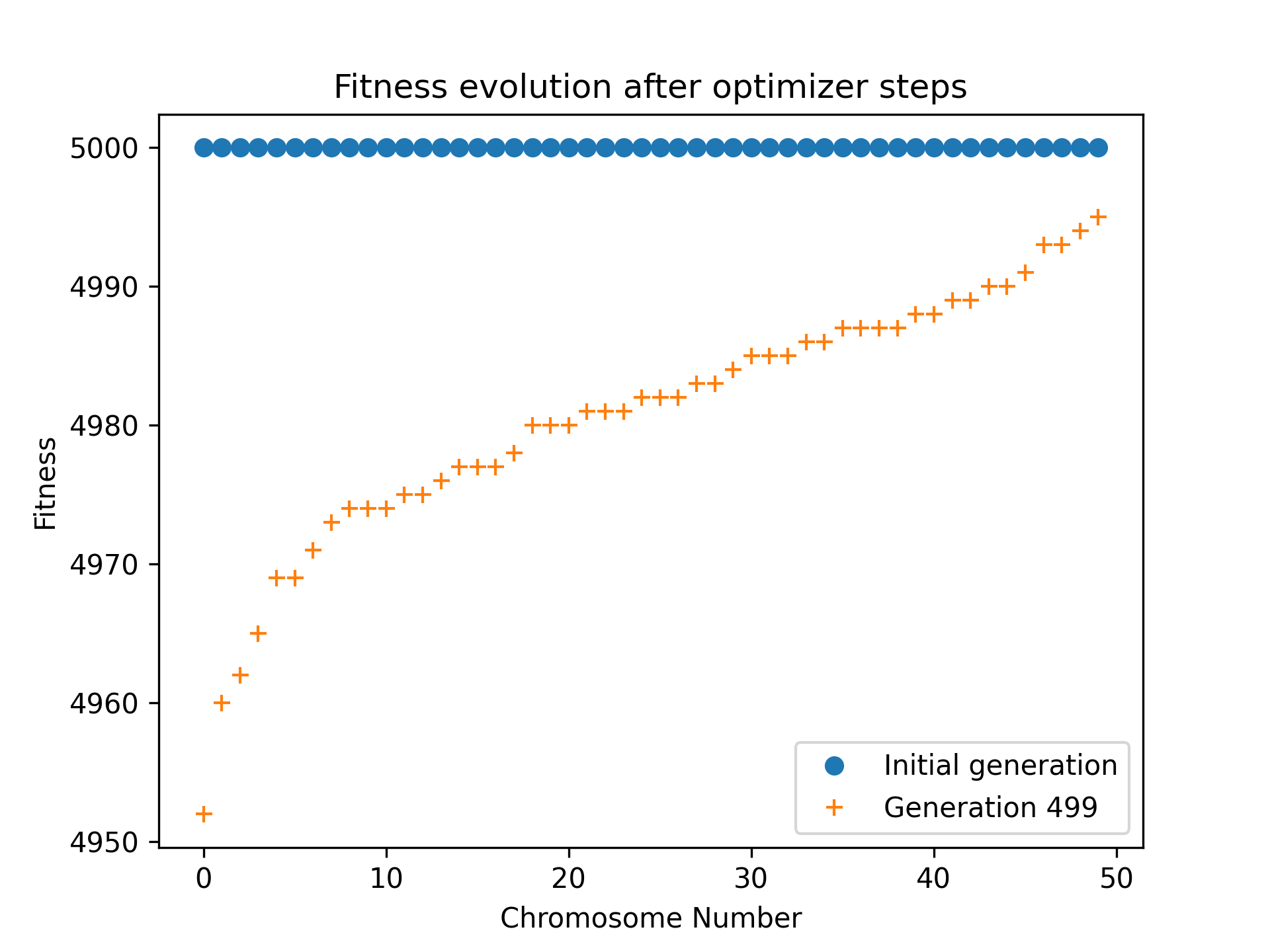}
\caption{Evolution of the fitness of the current population after 499 generations ($+$) versus the fitness of the initial population ($\bullet$) (see text).  The $+$ curve clearly shows the emergence of significantly more fit generation as a result of knowledge building from information provided by the student. The lowest fitness chromosome had a fitness of 4,952, and presented a final equation, whose use would yield a solution to the problem.}
\label{fig:fitness_evolve}
\end{figure}

Eventually, question possibilities were exhausted once again, and another step was taken by the GA.  With the collection of knowns having grown, the fitness values of some chromosomes should begin to decrease.  Indeed after many such iterations, the fitness evolved as shown in Fig.~\ref{fig:fitness_evolve}. Here the $+$ symbols clearly show a more fit population relative to the initial (random) generation.

\subsection{Performance}
As the population continued to evolve, the performance of the algorithm was tracked and is shown in Table~\ref{table:knowns}, which tabulates generations whose chromosomes led to questions that either elicited a direct response from the student (bold) or generated new knowledge based on accumulated data (underline).   Generation $130\dagger$ presents a final equation to the student, whose use will yield the answer to the problem. 

We note generation $1$ (entirely random) resulted in direct input of $2$ unique data items from the problem: the initial speed of the car at the traffic light and the time spent coasting. It also led to the identification of the two acceleration zones, enumerated as 0 (accelerating from the light) and 7 (coasting) by the chromosomes.

The algorithm was able to complete the equation ordering for the solution that we seek starting with the 3rd action of generation $11$.  Here, the position and speed of the car were found at the end of the time interval it spent accelerating away from the light.  This was followed by connecting these terminal parameters of that zone, to the initial parameters of the coasting zone in generations 16 and 20. The solution to the problem emerged as the second action of generation $130$\cite{spot_soln}.

\begin{center}
\begin{table}[h]
\centering
\begin{tabular}{ c | c | c | c | l }
Generation & Equation  & Variable & Zone & Response\\
\hline\hline
1 & 2 & $v_{0x}$ & 0 & {\bf 0 m/s} \\
1 & 1 & $v_{0x}$ & 0 & {\sl 0 m/s} \\
1 & 3 & $dt$ & 7 & {\bf 3 s} \\
1 & 1 & $dt$ & 7 & {\sl 3 s} \\
1 & 2 & $dt$ & 7 & {\sl 3 s} \\
4 & 1 & $a_x$ & 0 & {\bf 5 m/s$^2$} \\
4 & 2 & $a_x$ & 0 & {\sl 5 m/s$^2$} \\
9 & 1 & $x_0$ & 0 & {\bf 0 m} \\
11 & 3 & $dt$ & 0 & {\bf 8 s} \\
11 & 1 & $dt$ & 0 & {\sl 8 s} \\
11 & 2 & $dt$ & 0 & {\sl 8 s} \\
11 & 1 & $x$ & 0 & \underline{Solve for $x$} \\
11 & 2 & $v_x$ & 0 & \underline{Solve for $v_x$}  \\
16 & 1 & $x_0$ & 7 & $\leftarrow x$ from Zone 0 \\
20 & 2 & $v_{0x}$ & 7 & $\leftarrow v_x$ from Zone 0 \\
20 & 1 & $v_{0x}$ & 7 & $\leftarrow v_x$ from Zone 0 \\
130 & 2 & $a_x$ & 7 & {\bf 0 m/s$^2$} \\
130 & 1 & $a_x$ & 7 & 0 m/s$^2$ \\
130$\dagger$ & 1 & $x$ & 7 & \underline{Solve for $x$} \\
130 & 2 & $v_x$ & 7 & \underline{Solve for $v_x$}  \\
\hline\\
\end{tabular}
\caption{Performance of the algorithm for the sample problem shown above.  Generation $130\dagger$ represents the answer to the problem.  The table shows the generation number whose questions either elicited a response from the student or generated new knowledge based on accumulated data. Zone $7$ was found to be when the car was coasting, and $0$ when it was accelerating away from the traffic light.   Legend for ``Response'' column: {\bf Bold}: input directly from the user. {\sl Italic}: Internally generated as the given variable appears in multiple equations. \underline{Underline}: Internally generated using algebra.}
\label{table:knowns}
\end{table}
\end{center}

\begin{figure}[t]
\centering
\includegraphics[width=\columnwidth]{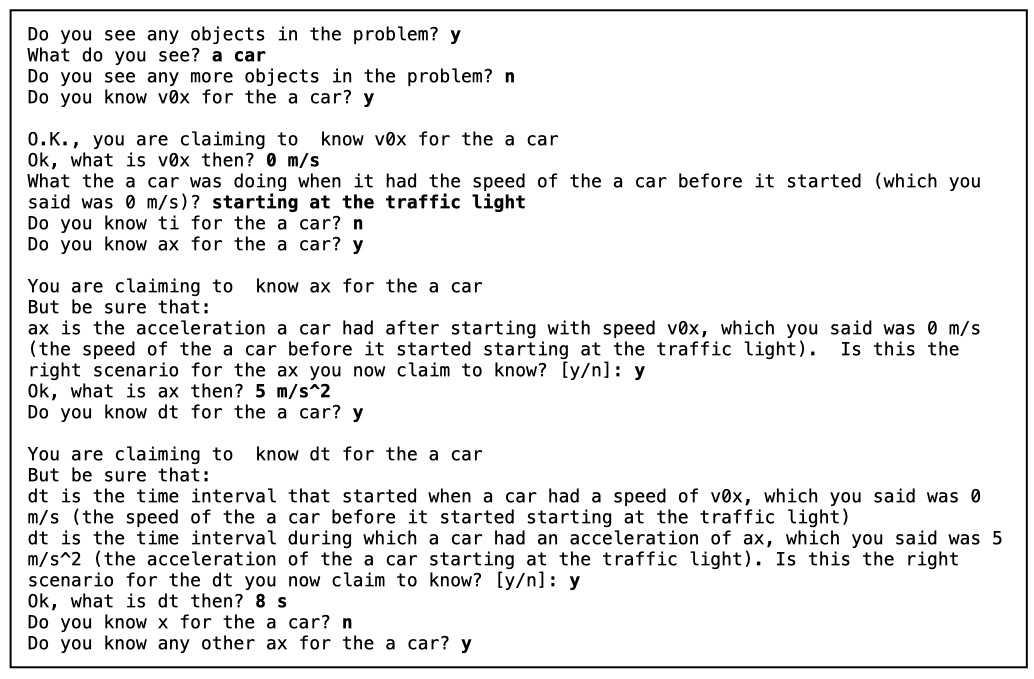}
\caption{A partial dialog with this algorithm at work for the example shown in Sect.~\ref{sect:example}.   Bold text was input from the user, plain text was generated by parsing the chromosomes as discussed.}
\label{fig:sample_run}
\end{figure}

\section{Interpretability}

This work carries some characteristics of artificial intelligence (AI), as seeing a computer pose questions and process answers may cause the computer to appear as an ``intelligent agent''\cite{norvig}. Thus there is a connection between the abstractness of a computer algorithm and the delivery of a personal result to a layperson, which begs the question of interpretability\cite{explain}\cite{xu}, which is a persistent difficulty with AI-type algorithms\cite{3b1b}. 

Given the well-known and conventional techniques for solving such physics problems and the small search space, we discuss some interpretability findings.

\subsection{Productivity}

We begin by defining {\sl productivity} of the algorithm, which is defined as the number of questions that elicited a response from the student.  We use the idea of ``productivity'' here since the algorithm caused the student to recognize something about their problem as motivated by the question itself.

\begin{figure}[t]
\centering
\includegraphics[width=\columnwidth]{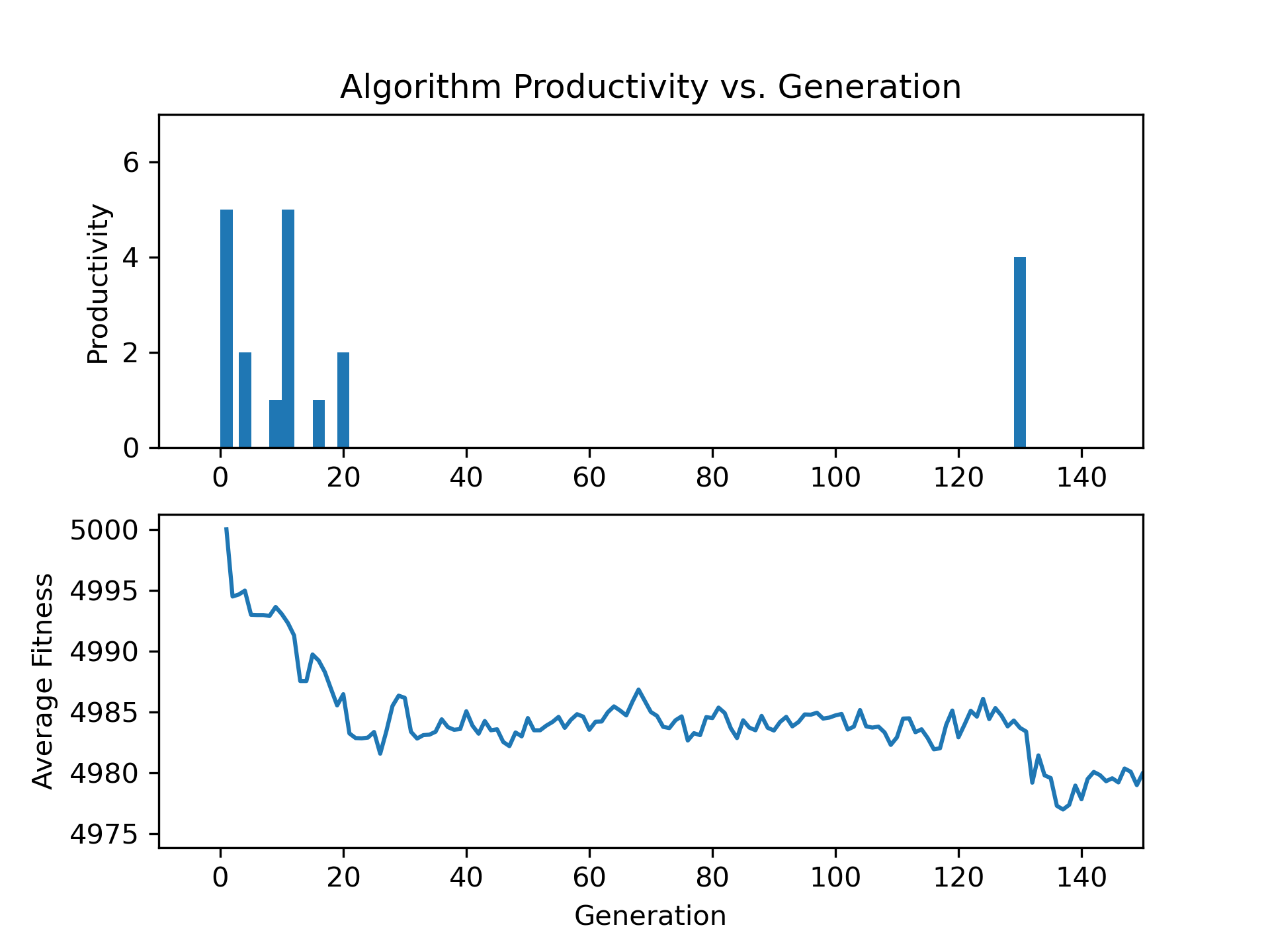}
\caption{Productivity of the algorithm (see text) and average population fitness as a function of generation.}
\label{fig:productivity_ga}
\end{figure}

Fig.~\ref{fig:productivity_ga} illustrates results for the problem stated above.  The lower plot shows the average fitness of the population at a given generation, which shows downward steps with each productive generation, indicating that unknowns for the equation set are decreasing as the algorithm gains input from the student. The small fluctuations are due to the continued population sorting and roulette wheel selection that produces each new generation (whether new data was found or not).

The upper plot shows the productivity, with the height of each bar indicating {\sl productivity}, or the number of responses gained from the student.  In this case, random generation $1$ was relatively productive, generating $2$ questions for which the student had a response. Then the algorithm continues to be very productive up to generation $20$.  At that point, with its optimization (i.e. crossover and mutations) still acting, it is unable to gain any input from the student until generation $130$, at which time a single  additional input was obtained, which was enough to solve the problem.

\begin{figure}[t]
\centering
\includegraphics[width=\columnwidth]{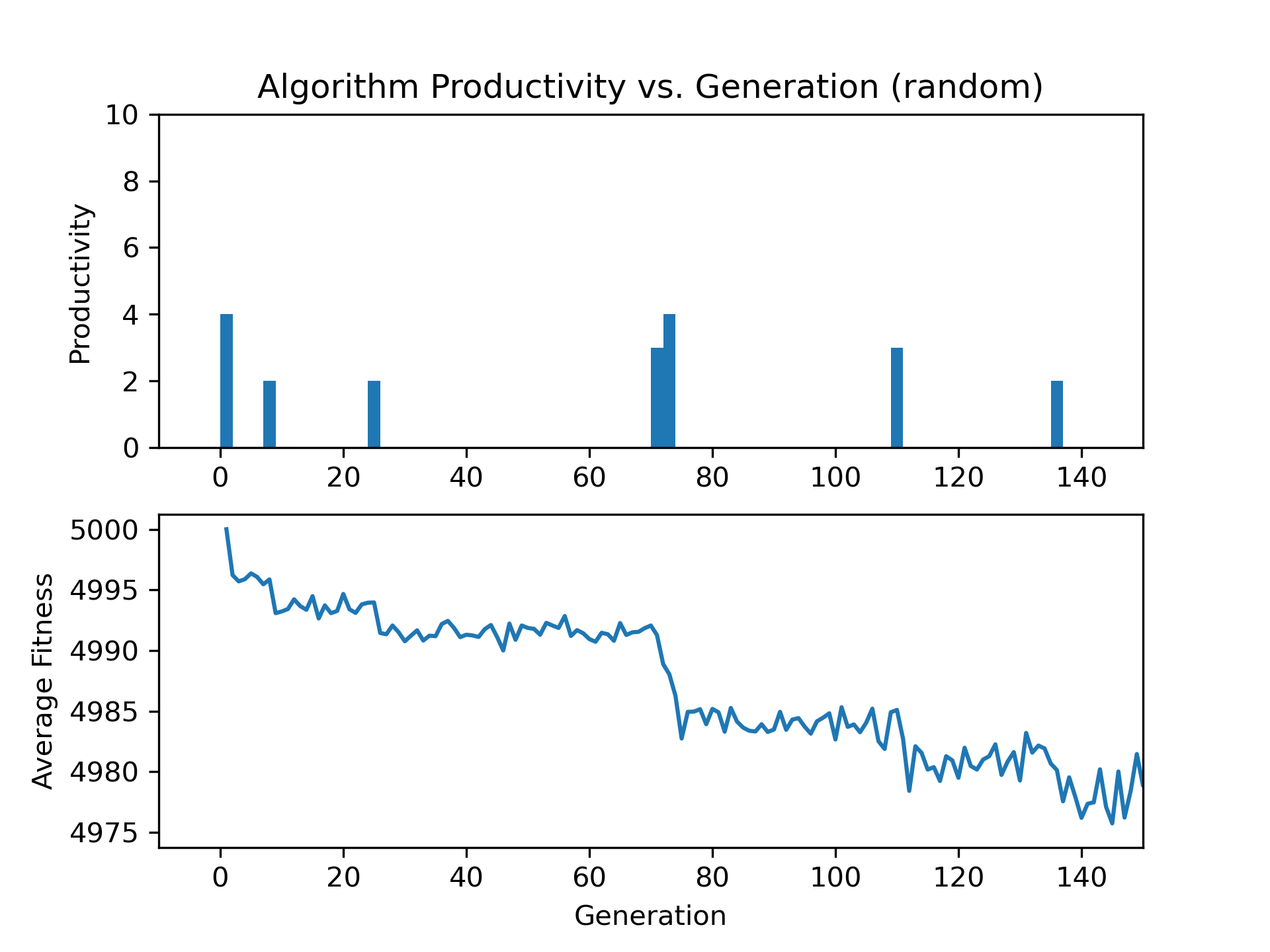}
\caption{Productivity of the algorithm (see text) and average population fitness as a function of generations that are entirely random relative to prior generations.}
\label{fig:productivity_rnd}
\end{figure}

As a control, we also ran the algorithm with questions generated from generations that were entirely random relative to the one prior.  We again measure the productivity as shown in Fig.~\ref{fig:productivity_rnd}. Here productive generations are more equally spaced as the algorithm proceeds, evidence of random chromosomes falling onto a question that elicits an answer by pure chance. There is little evidence of  large productivity groups  (i.e. clustered bars as in Fig.~\ref{fig:productivity_ga}), but also no long spans of zero productivity. The clustered bars in Fig.~\ref{fig:productivity_ga} are likely due to the GA's focus on asking questions about more fit equations given the roulette wheel selection used to generate future generations.  The randomly selected generations do not benefit from this technique.

This particular control run found the answer in generation $136$. Other tests have found a solution at  larger generation numbers, but never {\sl sooner} than the genetic algorithm.  The uniform productivity will likely not scale well with more complex problems\cite{more_complex}, leading to inefficient searches. 

The random control also reveals a lack of focus of the questions. It tends to noticeably bounce around amongst objects and equations, which may be confusing to the student. Solutions to physics problems should not be demonstrated as a random walk through available rules or constraints in a problem.

We know from the analytical solution that it is critical for the {\sl final} position and speed of the car as it leaves the acceleration zone to be connected to the {\sl initial} position and speed of the car as it enters the coast phase. With the genetic algorithm, Table~\ref{table:knowns} shows that the long unproductive run ended when the GA finally led to the connection between the final and initial positions in the first action of generation $130$. The solution followed immediately.  The random generations found this connection at generation $136$.

\subsection{Chromosome Evolution}

The small search space of this work makes interpretability findings driven by statistical properties of the chromosomes possible.  Here, the Diehard statistical tests for randomness\cite{diehard} were run on the chromosomes.

The initial (random) chromosomes pass all of the Diehard tests.  The minimum fitness chromosome, however, at the generation when the answer was found, also passes but shows a ``weak'' result for a unit-length serial test and lagged sums with lags of $15$ and $24$\cite{knuth}.

The weakened lagged sum of length $15$ and $24$ closely matches the tuple bit-length of $(12)$ and twice its length $(24)$\cite{diehard_more}. This may indicate that the genetic algorithm has introduced statistical correlations amongst bits in the chromosome with such lengths, as it works to find sequential bits in a chromosome that form productive tuples. Such  reordering will undoubtedly affect the distribution of 1s and 0s, as further indicated by the weakening serial test. 

We wouldn't expect either the crossover or mutation operation alone to alter the randomness of a chromosome. However, the selection bias of the roulette wheel selection for such operations may eventually cause this to happen, also affecting these tests.  Work in this area is ongoing.

\section{Conclusions}

In this work, we demonstrated an optimizer based on a genetic algorithm, that can guide a student to the solution of an introductory physics problem involving one-dimensional kinematics.  It does so by ordering equation use into a sequence that will lead to a solution. Questions about the problem are generated for the student directly from the chromosomes. Responses to these questions generate data that will decrease the number of unknowns in a domain of relevant equations. Equations with the fewest number of unknowns are considered the most fit  and can generate intermediate results that feed less fit equations.  This will eventually yield a sequence of equation use that can lead to the answer.

We also note a constraint this approach places on the GA. There is a general-to-specific trend in the tuples parsed from the chromosomes, meaning they start with knowledge of the existence of ``an object'' in a problem (which is very general). But for a solution, we must insist on knowing about the object all the way down to the particular acceleration zone that may experience, which is very specific. This refinement of the initially random population may have uses in other rule-based applications that require arriving at specific details despite starting with a very general idea.

We see other possible educational applications of this work as well. For example, this system could be used in geometry problems in finding parts of a right triangle or areas or volumes of various shapes given other quantities that may be known about the system.

We are also examining the prospects of routing textual output from the algorithm to an LLM to humanize the student-facing text generated by the algorithm. For example, the Python code will output message such as this one, as it grapples with the cautions discussed in Sect.~\ref{Pedagogical caution}:

\begin{quote}
{\sl
v0x is the speed of car at the beginning of the time interval dt, which you said was 8s (the time interval that the car spent went from stoplight)\\
\\
v0x is the speed of car it had as it just began to take on acceleration ax, which you said was 5 m/s\verb!^!2 (the acceleration of the car went from stoplight).
}
\end{quote}

\noindent
Asking an LLM (ChatGPT) to ``make this text more concise'' will result in 

\begin{quote}
{\sl
v0x is the car's speed at the start of the 8 s interval, when it began accelerating from the stoplight at 5 m/s\verb!^!2,
}
\end{quote}
which is much more desirable.  

A more systematic study of this approach for solving a variety of (more complex) problems in order to compile statistics on performance is planned. However, a testing backend must be developed as it is arduous to manually answer the algorithm's questions over and over again.  

Eventually, we plan on having an LLM play the role of the student, although the LLM performance in the similar environment of text-adventure games was mixed\cite{zork}. With human supervision, this may also lead to a method of evaluating LLM reliability in solving physics problems.  

Other work involves converting the codebase from Python to Javascript, so this algorithm may run in a web-browser for easy deployment to students.  Interpretability studies are ongoing.

\section{Acknowledgements}

The author (TB) wishes to thank Dr. M. Moelter for proofreading this work.  Thank you also to the four anonymous reviewers for the ``The Genetic and Evolutionary Computation Conference'' (GECCO 2025) whose input improved this paper considerably.  A special thank you also to the Evolutionary Computation and Machine Learning group at Victoria University in Wellington, New Zealand, for their kind and stimulating hospitality from Jan-April of 2024, where  this work began.

\pagebreak
\onecolumn
\appendix

Our algorithm, as it drives the GA  proceeds as follows. We use ideas from the Python language, as this was our implementation language of choice.

\begin{algorithm}
\caption{Complete logic of the GA-based algorithm presented here.}
\begin{algorithmic}[1]
\STATE $objects \leftarrow[\ ]$ (known objects)
\STATE $knowns \leftarrow[\ ]$ (known $(n,e,v,z)$ tuples)
\STATE $azones \leftarrow\{\ \}$ (known acceleration zones)
\STATE $equations \leftarrow$ [{'equation':'\verb!x=x0+v0 dt+0.5 a_x dt^2!','vars': [x,x0,v0,dt,a], 'var\verb$_$count': 5},...]
\STATE $var\_desc\leftarrow$ ['x': 'position of $<$object$>$ after it was done','x0': 'position of $<$object$>$ before it started',...]
\STATE $population \leftarrow$ ga\_init(size=50,length=12000)
\STATE $fitness\_list \leftarrow$ calc$\_$fitness($population$)
\STATE $generations\leftarrow 500$
\FOR{$g=1$ \TO $generations$}
	\FOR{chom=1 \TO 50}
		\FOR{$i=0$ \TO $12000-1$ {\bf step} 12}
			\STATE $(n,e,v,z) \leftarrow$  parsed from chromosome at index $i$
        			\IF{$n\notin$  objects}
				\STATE objects[$n$]$\leftarrow$prompt(``What do you see in the problem?'')
			\ENDIF
			\IF{$n\in$ objects \AND valid$(e,v$)}
				\IF{$(n,e,v)\notin $ knowns}
					\STATE yn$\_$var = prompt(``Do you know $v$ for object $n$?'')
					\IF{$z\notin$ azones}
						\STATE desc$\leftarrow$ prompt(``What was $n$ doing when it had $v$?'')
						\STATE azone$\leftarrow$ append($z$: desc)
					\ELSE
						\STATE yn$\leftarrow$ prompt(``Are you sure you mean $v$ while $n$ is doing $z$?'')
						\IF{yn == 'y' \AND yn$\_$var == 'y'}
							\STATE knowns$\leftarrow$ append([$n$,$v$,$e$,$z$])
						\ENDIF
					\ENDIF
				\ENDIF
				
				\# Look for the same variable in other equations
				\FOR{$eqn \in equations$ \AND $eqn \ne e$}
					\IF{$v\in eqn$}
						\STATE knowns$\leftarrow$ append([$n$,$v$,$eqn$,$z$])
					\ENDIF
				\ENDFOR

				\# Look for a possible variable to solve for
				\FOR{eqn in equations}
					\IF{known$\_$count(eqn) == variable$\_$count(eqn)-1}
						\STATE Solve for remaining unknown variable $v_{new}$
						\STATE  knowns$\leftarrow$ append([$n$,$v_{new}$,$e$,$z$])
						\STATE Message student about new ability to solve for $v_{new}$
					\ENDIF
				\ENDFOR
			\ENDIF \texttt{  }\# valid object
        		\ENDFOR \texttt{  }\# chromsome parsing
        	\ENDFOR \texttt{  }\# chromosome number
	\STATE $fitness\_list \leftarrow$ calc$\_$fitness($population$)
	\STATE Take a step with the GA (apply crossover and mutation operations)
\ENDFOR \texttt{  }\# generation
\end{algorithmic}
\end{algorithm}


\begin{thebibliography}{10}
\bibitem{coupon} We are guided here somewhat by the ``Coupon Collector's problem.'' In the sample problem, we have $1$ object, each with $12$ unknowns amongst the equation set. But we have two acceleration zones, making a total of $24$ total unknowns that need to be visited by some selection process. Strictly, we would need to throw $95$ random numbers before expecting each unknown to be selected for processing. Thinking of $95\approx 100$, our $1,000$ tuples are a $10\times$ ``safety'' factor to ensure the unknowns will be visited.
\bibitem{spot_soln} The algorithm is unaware of any particular solution goals, as its only job is the minimize unknowns in equations.  The solution needs to be spotted by the student.
\bibitem{more_complex} For example, we ran the algorithm for a problem involving the eventual meeting point of two runners.  One runner ran at a constant speed, and the other accelerated before also reaching a constant speed.  The genetic algorithm exhibited  productive sequences of generations early on, solving the problem in $387$ generations. The random algorithm needed $1,027$ generations and again exhibited evenly spaced productivity.
\bibitem{diehard_more} Lag tests up to $36$ or more, represented a group of three tuples is not available by the Diehard test suite.
\end{thebibliography}
\end{document}